%% file: main.tex
\documentclass[letterpaper, 10 pt, conference]{ieeeconf}  

\IEEEoverridecommandlockouts                              

\usepackage{times}

\usepackage{multicol}
\usepackage[bookmarks=true]{hyperref} 

\usepackage{times}
\usepackage{booktabs}
\usepackage{epsfig}
\usepackage{graphicx}
\usepackage{amsmath}
\usepackage{amssymb}
\usepackage{stfloats}\usepackage{multirow}
\usepackage{cite} 

\usepackage[noend]{algpseudocode}
\usepackage{algorithm}
\usepackage{xspace}
\usepackage{multirow}
\usepackage{microtype}

\DeclareMathAlphabet{\mathpzc}{OT1}{pzc}{m}{it}

\usepackage{color}
\newcommand{\red}[1]{\textcolor{red}{#1}}

\usepackage{todonotes} 
\usepackage{rotating}

\input{h2t_def.tex}

\title{\LARGE \bf
Introducing the Simulated Flying Shapes and Simulated Planar Manipulator Datasets }

\usepackage{authblk}

\author{Fabio Ferreira$^{\ast}{^\dagger}$, Jonas Rothfuss$^{\ast}{^\dagger}$, Eren Erdal Aksoy $^{\ast}{\ddagger}$, You Zhou$^{\ast}{\dagger}$ and Tamim Asfour$^{\dagger}$
\thanks{\ackTimestormREBA}
\thanks{$^{\dagger}$Institute for Anthropomatics and Robotics, Karlsruhe Institute of Technology, Karlsruhe, Germany.} 
\thanks{$^{\ddagger}$School of Information Technology, Halmstad University, Sweden}
\thanks{$^{\ast}$These authors contributed equally to this work.}
}

\begin{document}

\maketitle
\thispagestyle{empty}
\pagestyle{empty}

\begin{abstract}
We release two artificial datasets, \emph{Simulated Flying Shapes} and \emph{Simulated Planar Manipulator} that allow to test the learning ability of video processing systems. In particular, the dataset is meant as a tool which allows to easily   assess the sanity of deep neural network models that aim to encode, reconstruct or predict video frame sequences.
The datasets each consist of \num{90000} videos. The \emph{Simulated Flying Shapes} dataset comprises scenes showing two objects of equal shape (rectangle, triangle and circle) and size in which one object approaches its counterpart. The \emph{Simulated Planar Manipulator} shows a 3-DOF planar manipulator that executes a \emph{pick-and-place} task in which it has to place a size-varying circle on a squared platform. Different from other widely used datasets such as moving MNIST\footnote{\url{http://www.cs.toronto.edu/~nitish/unsupervised_video}} \cite{mnist,movingmnist}, the two presented datasets involve goal-oriented tasks (e.g. the manipulator grasping an object and placing it on a platform), rather than showing random movements. This makes our datasets more suitable for testing prediction capabilities and the learning of sophisticated motions by a machine learning model. This technical document aims at providing an introduction into the usage of both datasets. 
\end{abstract}


\input{flyingshapes.tex}

\input{planarmanipulator.tex}

\input{download.tex}

 

\bibliographystyle{IEEEtran}
\bibliography{references} 


\end{document}

%% file: h2t_def.tex
\usepackage[per=slash]{siunitx}
\usepackage{comment}

\definecolor{OliveGreen}{RGB}{0,200,25}

\newcommand{\ackTimestormREBA}{The research leading to these results has received  funding from the European Union's Horizon 2020 research and innovation programme under grant agreement No 641100 (TimeStorm) and from the German Research Foundation (DFG: Deutsche Forschungsgemeinschaft) under Priority Program on Autonomous Learning (SPP 1527).}

\newcommand{\replaced}[2]{\red{\ifmmode\text{\sout{\ensuremath{#1}}}\else\sout{#1}\fi}\darkgreen{#2}}
\newcommand{\removed}[1]{\red{\ifmmode\text{\sout{\ensuremath{#1}}}\else\sout{#1}\fi}}

 	\renewcommand{\replaced}[2]{#2}
	\renewcommand{\removed}[1]{}

\newcommand{\removedfootnote}[1]{\footnote{\removed{#1}}}
\newcommand{\removedsubsection}[1]{\subsection{\texorpdfstring{\removed{#1}}{#1}}}

	\renewcommand{\removedfootnote}[1]{}
	\renewcommand{\removedsubsection}[1]{}
	
	\renewcommand{\removedsubsection}[1]{}

%% file: flyingshapes.tex
\section{Simulated Flying Shapes}
The dataset consists of \num{90000} grayscale videos that show two objects of equal shape and size in which one object approaches the other one. The object speed during the process of approaching is hereby modeled by a proportional-derivative controller. Overall, three different shapes (rectangle, triangle and circle) are provided. Initial configurations of the objects such as the position or shape were randomly sampled. Different from the moving MNIST dataset, the samples comprise a goal-oriented task, namely one object has to fully cover the other object rather than randomly moving.

\begin{center}
\begin{figure}[ht]
	\includegraphics[width=0.43\textwidth]{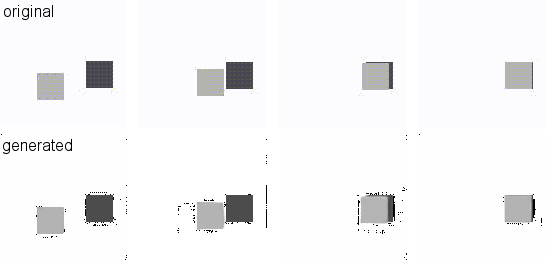}
	\caption{Simulated Flying Shapes example. Top: original data as publicly provided. Bottom: output generated by a deep auto-encoder which was trained on the Simulated Flying Shapes dataset. A moving example can be obtained from the repository.}
	\label{flyingshapes}
\end{figure}
\end{center}

For instance, one can use this artificial dataset as a testbed to investigate the capacity and output behavior of a deep neural network before testing it on real-world data. In a preceding research project we trained a deep auto-encoder network on both datasets. In figure \ref{flyingshapes} we show exemplary input and the corresponding output generated by the network introduced in \cite{DEM}.

We provide both the videos as .avi files as well as TensorFlow tfrecord files. Access to the files is provided via a GitHub repository\footnote{\url{https://github.com/ferreirafabio/FlyingShapesDataset}}. The samples in the tfrecord files contain 10 frames of each original video which were taken equally distributed over their entire playtime. Additional technical information are provided in the following subsections.

\subsection{Video File Specifications}
The .avi video files have the following specifications:
\begin{itemize}
\item \texttt{video resolution}: 128$\times$128
\item \texttt{fps}: 30
\item \texttt{color depth}: 24bpp (3 channels, grayscale)
\item \texttt{video codec}: ffmjpeg
\item \texttt{compression format}: mjpeg
\item \texttt{color encoding}: yuvj420p
\item the samples follow the naming:\\
\texttt{id\_shape\_startLocation\_endLocation\_\\motionDirection\_euclideanDistance}
\item where:
\begin{itemize}
	\item \texttt{id} is a unique identifier
	\item \texttt{startLocation} starting position of the object, e.g. righttop
	\item \texttt{endLocation}: destination position of the object, e.g. leftbottom
	\item \texttt{motionDirection}: motion direction of moving object, e.g. left
	\item \texttt{euclideanDistance}: Euclidean distance between the two objects, e.g. 7.765617
	\end{itemize}
\end{itemize}

\subsection{tfrecord Specifications}
The tfrecord files have been created with the pip package \texttt{video2tfrecord} and each file contains 1000 videos.
Due to the high computational cost of processing all original video frames in deep neural networks, we decided to reduce the number of extracted frames for the tfrecord files. As a result, every single tfrecord file entry consists of 10 RGB frames which were taken equally distributed over the video playtime. Assuming no prior knowledge about the video and its inherent scene dynamics, choosing frames equally spaced maximizes the chances of capturing most of the spatio-temporal dynamics. The files store both the videos itself and meta information from the file name (start location, eucl. distance etc.). The video data is stored in a \emph{feature} dict which is serialized as tf.train.Example and contains the following keys:
\begin{itemize}
\item \texttt{feature}['path'] 
\item \texttt{feature}['height']
\item \texttt{feature}['width']
\item \texttt{feature}['depth']
\item \texttt{feature}['id']
\end{itemize}
Additional information is stored in a dictionary \emph{meta\_dict} which is also serialized within the \emph{feature} dict, accessible by the key \emph{metadata}. It contains the following additional keys:
\begin{itemize}
\item \texttt{meta\_dict}['start\_location']
\item \texttt{meta\_dict}['end\_location']
\item \texttt{meta\_dict}['motion\_location']
\item \texttt{meta\_dict}['eucl\_distance']
\end{itemize}

%% file: planarmanipulator.tex
\section{Simulated Planar Manipulator}
\begin{center}
\begin{figure}[ht]
	\includegraphics[width=0.43\textwidth]{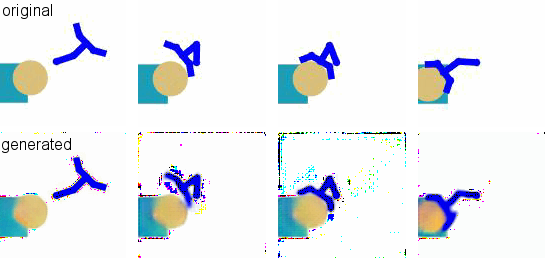}
	\caption{Simulated Flying Shapes example. Top: original data as publicly provided. Bottom: output generated by a deep auto-encoder which was trained on the Simulated Flying Shapes dataset. A moving example can be obtained from the repository (see attached URL).}
	\label{planarmanipulator}
\end{figure}
\end{center}

The dataset consists of \num{90000} color videos that show a planar robot manipulator executing articulated manipulation tasks. More precisely, the manipulator grasps a circular object of random color and size and places it on top of a square object/platform of again random color and size. The initial conﬁgurations (location, size and color) of the objects were randomly sampled during generation. Similarly to the Flying Shapes dataset, the samples again comprise a goal-oriented task as described above, making it highly suitable for testing prediction capabilities and sanity-checks of ML models. In figure \ref{planarmanipulator} we show exemplary input and the corresponding output, again generated by the network architecture used in \cite{DEM}.

Both videos as .avi files as well as TensorFlow tfrecord files can be accessed via our GitHub repository\footnote{\url{https://github.com/ferreirafabio/PlanarManipulatorDataset}}. Similar to the Flying Shapes dataset, the tfrecord files contain 10 frames of each original video, again taken equally distributed over their playtime. Technical information can be extracted from the following subsections.

\subsection{Video File Specifications}

The .avi video files have the following specifications:
\begin{itemize}
\item \texttt{video resolution}: 128$\times$128
\item \texttt{fps}: 30
\item \texttt{color depth}: 24bpp (3 channels, RGB)
\item \texttt{video codec}: ffmjpeg
\item \texttt{compression format}: mjpeg
\item \texttt{color encoding}: yuvj420p 
\end{itemize}

\subsection{tfrecord Specifications}
The tfrecord files have been created with the pip package \texttt{video2tfrecord} and each file contains \num{1000} videos. For the same reasons as for the Simulated Flying Shapes dataset, every single tfrecord file entry consists of 10 RGB frames which were taken equally distributed over the video playtime. Again the video data is stored in a \emph{feature} dict which is serialized as tf.train.Example and contains the following keys:
\begin{itemize}
\item \texttt{feature}['path'] 
\item \texttt{feature}['height']
\item \texttt{feature}['width']
\item \texttt{feature}['depth']
\item \texttt{feature}['id']
\end{itemize}
In contrast to the Simulated Flying Shapes dataset, the tfrecord meta dict does not have additional information related to the video content.

%% file: download.tex
\section{Download}
Videos and tfrecords of both datasets are provided as \texttt{.tar.gz} chunk files which can be downloaded in two ways:
\begin{itemize}
\item use the download scripts \texttt{download\_videos.py} or \texttt{download\_tfrecords.py} provided in the respective repositories and run them with Python to download them directly into the script directory.
\sloppy
\item open the \texttt{\{flyingshapes,}\texttt{planarmanipulator\}}\\\texttt{\_videos.txt} file (or \texttt{\_tfrecords.txt}) provided in the repositories and use the links to directly download the files with a browser.
\end{itemize}

%% file: main.bbl
\begin{thebibliography}{1}
\providecommand{\url}[1]{#1}
\csname url@samestyle\endcsname
\providecommand{\newblock}{\relax}
\providecommand{\bibinfo}[2]{#2}
\providecommand{\BIBentrySTDinterwordspacing}{\spaceskip=0pt\relax}
\providecommand{\BIBentryALTinterwordstretchfactor}{4}
\providecommand{\BIBentryALTinterwordspacing}{\spaceskip=\fontdimen2\font plus
\BIBentryALTinterwordstretchfactor\fontdimen3\font minus
  \fontdimen4\font\relax}
\providecommand{\BIBforeignlanguage}[2]{{%
\expandafter\ifx\csname l@#1\endcsname\relax
\typeout{** WARNING: IEEEtran.bst: No hyphenation pattern has been}%
\typeout{** loaded for the language `#1'. Using the pattern for}%
\typeout{** the default language instead.}%
\else
\language=\csname l@#1\endcsname
\fi
#2}}
\providecommand{\BIBdecl}{\relax}
\BIBdecl

\bibitem{mnist}
Y.~Lecun, L.~Bottou, Y.~Bengio, and P.~Haffner, ``Gradient-based learning
  applied to document recognition,'' \emph{Proceedings of the IEEE}, vol.~86,
  no.~11, pp. 2278--2324, Nov 1998.

\bibitem{movingmnist}
\BIBentryALTinterwordspacing
N.~Srivastava, E.~Mansimov, and R.~Salakhutdinov, ``Unsupervised learning of
  video representations using lstms,'' \emph{CoRR}, vol. abs/1502.04681, 2015.
  [Online]. Available: \url{http://arxiv.org/abs/1502.04681}
\BIBentrySTDinterwordspacing

\bibitem{DEM}
J.~Rothfuss, F.~Ferreira, E.~E. Aksoy, Y.~Zhou, and T.~Asfour, ``Deep episodic
  memory: Encoding, recalling, and predicting episodic experiences for robot
  action execution,'' 2018.

\end{thebibliography}


@ARTICLE{mnist, 
author={Y. Lecun and L. Bottou and Y. Bengio and P. Haffner}, 
journal={Proceedings of the IEEE}, 
title={Gradient-based learning applied to document recognition}, 
year={1998}, 
volume={86}, 
number={11}, 
pages={2278-2324}, 
keywords={backpropagation;convolution;multilayer perceptrons;optical character recognition;2D shape variability;GTN;back-propagation;cheque reading;complex decision surface synthesis;convolutional neural network character recognizers;document recognition;document recognition systems;field extraction;gradient based learning technique;gradient-based learning;graph transformer networks;handwritten character recognition;handwritten digit recognition task;high-dimensional patterns;language modeling;multilayer neural networks;multimodule systems;performance measure minimization;segmentation recognition;Character recognition;Feature extraction;Hidden Markov models;Machine learning;Multi-layer neural network;Neural networks;Optical character recognition software;Optical computing;Pattern recognition;Principal component analysis}, 
doi={10.1109/5.726791}, 
ISSN={0018-9219}, 
month={Nov},}


@article{movingmnist,
  author    = {Nitish Srivastava and
               Elman Mansimov and
               Ruslan Salakhutdinov},
  title     = {Unsupervised Learning of Video Representations using LSTMs},
  journal   = {CoRR},
  volume    = {abs/1502.04681},
  year      = {2015},
  url       = {http://arxiv.org/abs/1502.04681},
  archivePrefix = {arXiv},
  eprint    = {1502.04681},
  timestamp = {Wed, 07 Jun 2017 14:41:19 +0200},
  biburl    = {https://dblp.org/rec/bib/journals/corr/SrivastavaMS15},
  bibsource = {dblp computer science bibliography, https://dblp.org}
}



@misc{DEM,
Author = {Jonas Rothfuss and Fabio Ferreira and Eren Erdal Aksoy and You Zhou and Tamim Asfour},
Title = {Deep Episodic Memory: Encoding, Recalling, and Predicting Episodic Experiences for Robot Action Execution},
Year = {2018},
Eprint = {arXiv:1801.04134},
}
